\documentclass[10pt,twocolumn,letterpaper]{article}

\usepackage{cvpr}              

\usepackage{lineno}

\usepackage{multirow} 
\usepackage{makecell}
\usepackage{color, colortbl}
\definecolor{mylb}{RGB}{229, 247, 255}
\newcommand{\br}{\rowcolor[RGB]{229, 247, 255}}
\definecolor{cvprblue}{rgb}{0.21,0.49,0.74}
\usepackage[pagebackref,breaklinks,colorlinks,allcolors=cvprblue]{hyperref}
\newcommand\blfootnote[1]{%
  \begingroup
  \renewcommand\thefootnote{}\footnote{#1}%
  \addtocounter{footnote}{-1}%
  \endgroup
}
\def\paperID{4628} 
\def\confName{CVPR}
\def\confYear{2025}

\title{DVHGNN: Multi-Scale Dilated Vision HGNN for Efficient Vision Recognition}

\author{Caoshuo Li\textsuperscript{1,2}$^*$, Tanzhe Li\textsuperscript{1,2}$^*$, Xiaobin Hu\textsuperscript{3}, Donghao Luo\textsuperscript{3}, Taisong Jin\textsuperscript{1,2}$^\dag$\\
\textsuperscript{1}Key Laboratory of Multimedia Trusted Perception and Efficient Computing, \\
Ministry of Education of China, Xiamen University, China.\\ 
\textsuperscript{2}School of Informatics, Xiamen University, China. \textsuperscript{3}Tencent Youtu Lab. \\
{\tt\small \{licaoshuo, tanzheli\}@stu.xmu.edu.cn, jintaisong@xmu.edu.cn,} \\
{\tt\small \{xiaobinhu, michaelluo\}@tencent.com}
}

\begin{document}
\maketitle
\begin{abstract}
Recently, Vision Graph Neural Network (ViG) has gained considerable attention in computer vision. Despite its groundbreaking innovation, Vision Graph Neural Network encounters key issues including the quadratic computational complexity caused by its K-Nearest Neighbor (KNN) graph construction and the limitation of pairwise relations of normal graphs. To address the aforementioned challenges, we propose a novel vision architecture, termed \textbf{D}ilated \textbf{V}ision \textbf{H}yper\textbf{G}raph \textbf{N}eural \textbf{N}etwork (DVHGNN), which is designed to leverage multi-scale hypergraph to  \emph {efficiently} capture high-order correlations among objects. Specifically, the proposed method tailors Clustering and \textbf{D}ilated \textbf{H}yper\textbf{G}raph \textbf{C}onstruction (DHGC) to adaptively capture multi-scale dependencies among the data samples. Furthermore,  a dynamic hypergraph convolution mechanism is proposed to facilitate adaptive feature exchange and fusion at the hypergraph level. Extensive qualitative and quantitative evaluations of the benchmark image datasets demonstrate that the proposed DVHGNN significantly outperforms the state-of-the-art vision backbones. For instance, our DVHGNN-S achieves an impressive top-1 accuracy of \textbf{83.1\%}  on ImageNet-1K, surpassing ViG-S by \textbf{+1.0}$\uparrow$ and ViHGNN-S by \textbf{+0.6}$\uparrow$.
\blfootnote{$^*$Equal contribution  \quad $^\dag$Corresponding author }
\end{abstract}    
\section{Introduction}
\label{sec:intro}

\begin{figure}
    \centering
    \includegraphics[width=1.0\linewidth]{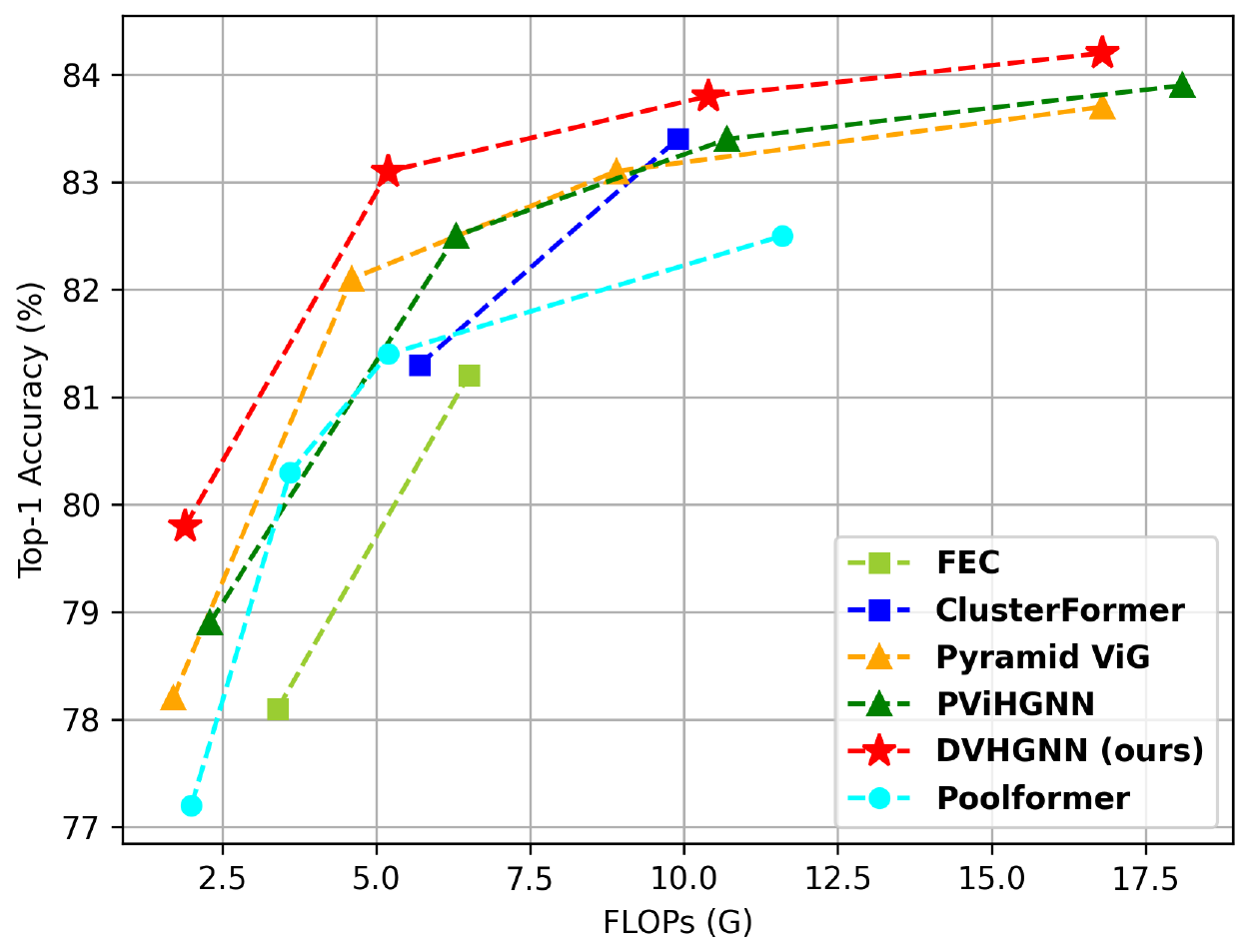}
    \caption{Comparison of FLOPs and Top-1 accuracy on ImageNet-1K. The proposed DVHGNN achieves the best performance compared to other state-of-the-art models.}
    \label{fig:enter-label}
\end{figure}
The rapid advancement of deep learning has significantly propelled computer vision community. Convolutional Neural Networks (CNNs)~\cite{alexnet,googlenet, vgg, resnet, efficientnet, depsconv,dcn} have become the predominant approach in various vision tasks, efficiently capturing the spatial relationships and structural complexities within images owing to the locality and shared weights. However, CNNs are constrained by their narrow focus on local information, rendering them incapable of capturing the long-range dependencies. The introduction of the Transformer~\cite{transfomer}, renowned for handling long-range information in natural language processing (NLP), to computer vision via the Vision Transformer (ViT)~\cite{vit} marked a significant shift. ViT demonstrates strong performance across various vision tasks by efficiently modeling long-range dependencies of an image. Despite its strengths, ViT still has the drawbacks including the absence of CNNs' inductive bias, a dependency on extensive training data and the quadratic computational complexity caused by its global self-attention mechanism. Subsequently, various variants and extensions~\cite{deit,pvt,swin,dat,SlideTransformer,clusterformer} of ViT have been proposed to mitigate these drawbacks, offering more efficient and adaptable solutions for different vision tasks.

Both CNNs and ViTs treat images as grid and sequential structures, respectively, lacking the flexibility and the ability to capture the structure of complex objects in an image. Recently, Vision Graph neural network (ViG)~\cite{vig} transforms images into graph structures within non-Euclidean space and leverages Graph Neural Networks (GNNs) to employ feature exchange and fusion at the graph level. This marks the first successful generalization of GNNs to a general vision backbone, bridging the gap in capturing complex object structures with enhanced flexibility.

However, ViG still faces two key limitations. \emph {On the one hand}, despite efforts to extract deeper semantic information through an increased number of blocks, ViG fails to represent the complex inter-class and intra-class relationships among objects in an image effectively. This limitation stems from its normal graph structure, designed to modeling the connections of low-level and local features between neighboring nodes that fails to capture the high-order correlations among more than two nodes.
\emph {On the other hand},  the employment of the KNN graph in ViG has quadratic computational complexity, which requires a substantial amount of memory and computational resources. Furthermore, the inherent non-learnability of the KNN graph strategy results in the potential information lost in the graph construction.

To overcome the abovementioned issues, models like ViHGNN~\cite{vihgnn} are proposed to enhance ViG's capability via hypergraph capturing high-order correlations among objects. However, ViHGNN shows marginal performance improvement based on  two reasons: 

\noindent{\textbf{1) Limitations of Hypergraph Construction: }}ViHGNN employs the Fuzzy C-means algorithm to construct hypergraphs across the entire image, neglecting the significance of local and multi-scale features, resulting in quadratic computational complexity for the number of hyperedges.

\noindent{\textbf{2) Reciprocal Feedback Limitations: }}Although ViHGNN introduces a reciprocal feedback mechanism between patch embeddings and the hypergraph for message passing and mutual optimization, the nature of the Fuzzy C-means still constrains the model's adaptability during the learning process. Specifically, ViHGNN lacks the capability to effectively model the  dynamic connections between hypergraph structure and hypergraph convolution.

To address the issues above, we propose a novel Vision Hypergraph Neural Network, termed \textbf{D}ilated \textbf{V}ision \textbf{H}yper\textbf{G}raph \textbf{N}eural \textbf{N}etwork (DVHGNN), which introduces a multi-scale hypergraph approach to image representation. Specifically, a multi-scale hypergraph is constructed using cosine similarity clustering and Dilated HyperGraph Construction (DHGC) techniques. In this process, each hyperedge generates a centroid, allowing each vertex to adaptively perform dynamic hypergraph convolution by utilizing either cosine similarity or sparsity-aware weights relative to the centroid. Our contributions are summarized as follows:%
\begin{itemize}[topsep=3pt, partopsep=3pt,leftmargin=15pt, itemsep=3pt]
    \item We propose a novel paradigm termed DVHGNN, which is designed to leverage multi-scale hypergraph to efficiently capture high-order correlations among objects.
    \item To obtain multi-scale hypergraph representations of the images, we adopted a dual-path design that includes clustering and DHGC, simultaneously focusing on local and sparse multi-scale information. 

    \item To better facilitate information exchange at the hypergraph level, we propose a novel two-stage Dynamic Hypergraph Convolution framework, which leverages the pairwise cosine similarity and sparsity-aware weight of hyperedges to adaptively aggregate vertex embeddings into hyperedge features to update original embeddings.

    \item Extensive experiments demonstrate the superior performance of our DVHGNN. Specifically, our DVHGNN-S achieves an impressive Top-1 accuracy of 83.1\% on ImageNet-1K, surpassing ViG-S by 1.4\% and ViHGNN-S by 0.6\% with fewer parameters and FLOPs.
   
\end{itemize}

\section{Realted Works}
\label{sec:relatedwork}

\subsection{CNNs and Transformer for Vision}
Convolutional Neural Networks (CNNs)~\cite{cnn} have been the cornerstone of computer vision, since AlexNet~\cite{alexnet} marked a significant breakthrough, leading to the development of various influential CNN architectures including VGG~\cite{vgg}, GoogleNet~\cite{googlenet}, ResNet~\cite{resnet}, and MobileNet~\cite{depsconv}. Inspired by the success of Transformer~\cite{transfomer} in natural language processing, Vision Transformer~\cite{vit} was introduced into  vision tasks, which is used as the basis of the subsequent models like DeiT~\cite{deit}, PVT~\cite{pvt}, and Swin-Transformer~\cite{swin} to achieve impressive performances across various downstream vision tasks. Furthermore, hybrid architectures such as CvT~\cite{cvt}, Coatnet~\cite{coatnet}, and ViTAE~\cite{vitae} have been proposed to amalgamate the strengths of convolutions and Transformers. Recently, larger kernel CNNs, such as ConvNeXt~\cite{convnet} and RepLKNet~\cite{RepLKNet}, have demonstrated strong competitive capabilities, highlighting the diverse and evolving landscape in advancing computer vision technologies. However, CNNs still struggle with long-range modeling.
\begin{figure*}[t]
    \centering
    \includegraphics[width=1.0\linewidth]{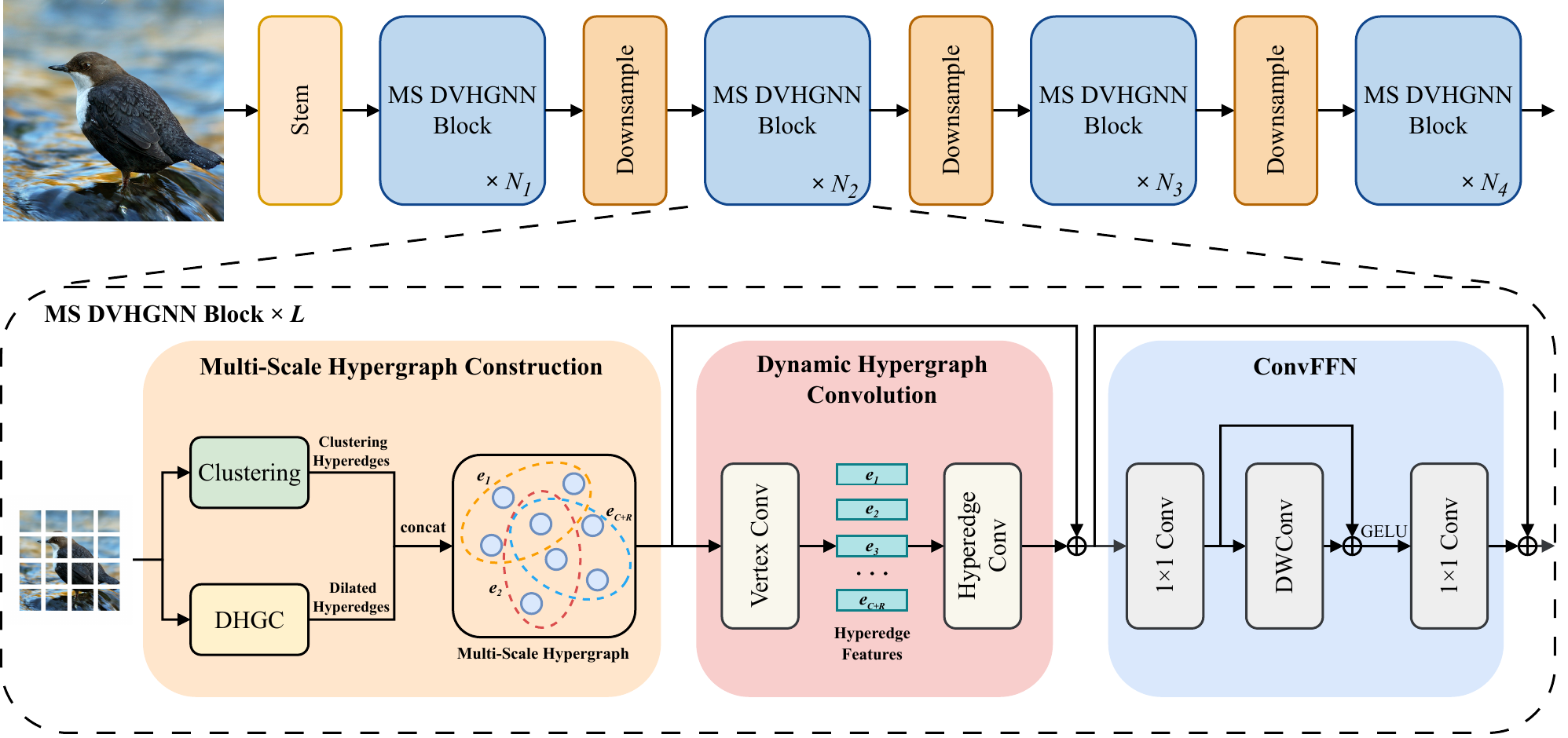}
    \caption{Architecture of the proposed DVHGNN. In each block, Multi-Scale(MS) DVHGNN block constructs multi-scale hyperedges, followed by message passing through vertex and hyperedge convolutions, and finalizes with ConvFFN to enhance feature transformation capacity and counteract over-smoothing.}
    \label{fig:attn_comparison}
\end{figure*}
\subsection{Graph/Hypergraph Neural Network}
Graph Neural Networks (GNNs)~\cite{gnn,gnn1} were initially used for processing non-Euclidean data and have achieved various applications in social networks~\cite{social_network}, citation networks~\cite{citation_network}, and biochemical graphs~\cite{Chemical_graph}. Currently, GNNs have been applied to computer vision, including object detection~\cite{detector1}, and point cloud classification~\cite{pointclassification}.

In computer vision, HGNNs are regarded as a significant extension of GNNs, distinguished by their ability to capture high-order relationships through hypergraphs. This shift enhances the understanding and representation of complex data structures, as demonstrated in applications such as image retrieval~\cite{hyper_image_retrieval}, 3D object classification~\cite{hyper_3-d_object_retrieval, hyper_video_seg}, and person re-identification~\cite{person}. Specifically, HGNNs improve image retrieval by modeling images as vertices connected by hyperedges based on feature correlations, thereby refining the retrieval process with greater sensitivity to image relationships. Similarly, in 3D object classification, HGNNs enhance accuracy by leveraging vertices for object representation and hyperedges to capture the complex relationships between different views. Despite the contributions made by GNNs and HGNNs, there has been limited use of GNNs or HGNNs as the backbone in computer vision. ViG~\cite{vig}  and ViHGNN~\cite{vihgnn} are among the few works that treat patches within input images as nodes and construct graphs/hypergraphs through different strategies like KNN or Fuzzy C-means. However, both KNN and Fuzzy C-means face challenges in high-dimensional spaces, suffer from computational inefficiency, and struggle to capture complex patterns, making them less suitable for modern visual tasks compared to more adaptive deep learning models.

\subsection{Graph/Hypergraph Structure Learning}
In GNNs and HGNNs, the performance is significantly influenced by the quality of graph and hypergraph structures. Recent advances have focused on structure learning to enhance the effectiveness of these models. For GNNs, methods such as those in~\cite{graph_learn_1, graph_learn_2} have been developed to jointly learn graph structures and node embeddings, relying on graph adjacency matrix properties like low-rank and sparsity. Other approaches~\cite{graph_learn_3, graph_learn_4} dynamically adjust graph structures by learning metrics or distributions for the edges.

Hypergraph structure learning, while less explored, has seen advances with models like DHSL~\cite{dhsl_1, dhsl_2} applying dual optimization for simultaneous learning of label projection matrices and hypergraph structures. DHGNN~\cite{dhgnn} uses K-Means and KNN for adaptive hypergraph construction, and HERALD~\cite{laphl} focuses on optimizing hypergraph Laplacian matrices. Challenges include non-convex optimization issues and high computational demands, especially noted in DHSL.
Expanding the repertoire of structure learning, recent works include HyperSAGE~\cite{hypersage} for scalable hypergraph learning and Dynamic HyperEdge Convolution Networks (DHECN) that adjust hyperedge weights dynamically. However, these methods are constrained by their graph construction techniques, preventing them from adaptively adjusting their hypergraph structures.
\section{Methods}

In this section, we begin by revisiting the concept of hypergraph. Subsequently, we delineate the process of the proposed multi-scale hypergraph image representation. Finally, we elaborate on the design of the dynamic hypergraph convolution and the multi-head computation mechanism.

\subsection{Preliminary:Recap the Concept of Hypergraph}
\noindent{\textbf{Notations. }}For an image with dimensions  $ H \times W \times 3 $, we employ  Vision GNN~\cite{vig} to divide it into $ N $ patches. A linear layer is then applied to the derived patches to convert them into high-dimensional vectors $ \mathbf{x}_i \in \mathbb{R}^D $. This transformation results in a matrix $ \mathbf{X} = \{\mathbf{x}_1, \mathbf{x}_2, \ldots, \mathbf{x}_N\} $, where $ D $ represents the dimensionality of the features and $ i $ indexes the patches with $ i = 1, 2, \ldots, N $. These vectors can be conceptualized as a collection of unordered vertices, denoted by $ \mathcal{V} = \{v_1, v_2, \ldots, v_N\} $. A hypergraph is defined as $\mathcal{G} = (\mathcal{V}, \mathcal{E}, \mathbf{W})$, where $\mathcal{V}$ is a set of vertices and $\mathcal{E}$ is a set of hyperedges. The set of incident edges of vertex $i$ is denoted by $E_i$, where $E_i = \{e \in E \mid v_i \in e\}$. 
The weight of each hyperedge is encoded by 
$\mathbf{W}$, a diagonal matrix of edge weights.  Differing from a normal graph, each hyperedge can connect any number of vertices. 


\begin{figure}[t]
  \centering
    \includegraphics[width=\linewidth]{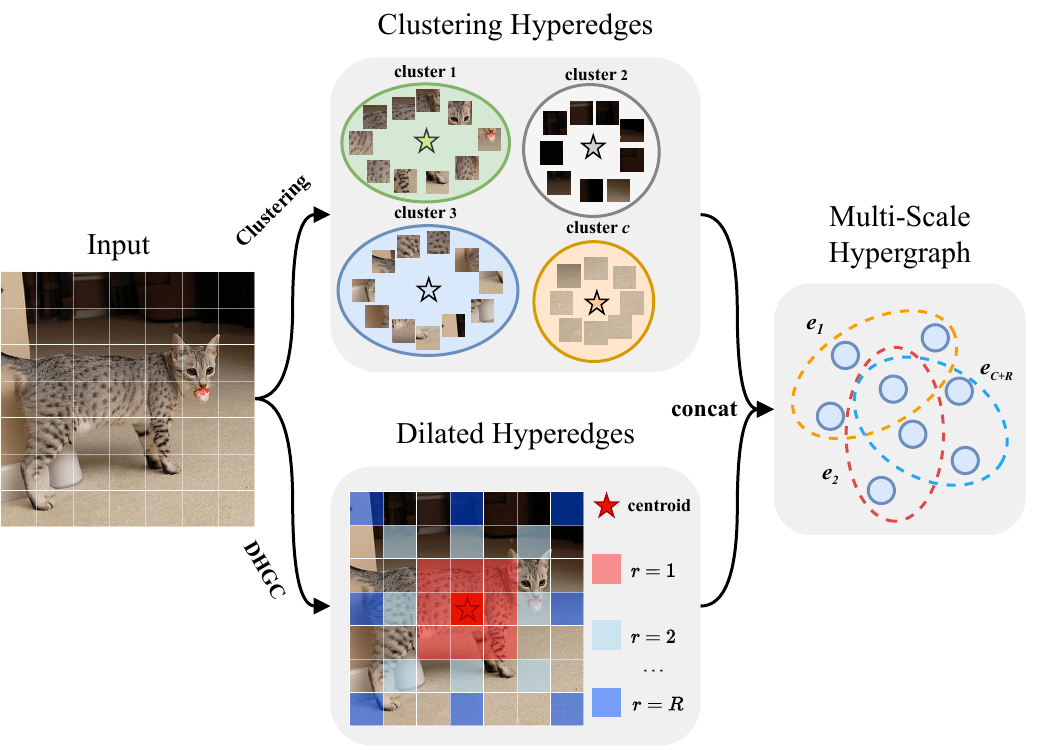}
  \caption{Illustration of Multi-Scale Hypergraph Construction (without region partition). The final hyperedge set is composed of two types of hyperedges: a set of size $C$ obtained from cosine similarity clustering, and a set of size $R$ derived from DHGC. Each hyperedge corresponds to a hyperedge centroid, marked with a pentagon in the diagram. By default, $R$ = 3, with the distinct dilated hyperedges corresponding to a kernel size of 3 $\times$ 3 with dilation rates $r$ = 1, 2, and 3, respectively, resulting in receptive field sizes of 3 $\times$ 3 ,  5 $\times$ 5 , and 7 $\times$ 7 .}
  \label{fig:ms hypergraph}
\end{figure} %

\subsection{Multi-Scale HGNN Representation of Image}
In this subsection, we propose the Cluster and Dilated Hypergraph Construction (DHGC) based
Hypergraph representation. Our hypergraph construction approach includes two distinctive hyperedge types: ones derived from clustering and the others obtained via DHGC. Initially, a hyperedge set is generated through a clustering technique. The derived set is then augmented by employing a DHGC mechanism, culminating in a multi-scale hyperedge collection characterized by its sparsity.
The details of our Multi-Scale Hypergraph Construction are shown in Figure.~\ref{fig:ms hypergraph}.

\noindent{\textbf{Clustering. }}Given a set of vertices $\mathcal{V}$ and associated feature vectors $\mathbf{X} \in \mathbb{R}^{N \times D}$, vertices are grouped into the distinct clusters reflecting their mutual similarities and ensuring that each vertex belongs exclusively to a single cluster. Initially, the feature vectors $\mathbf{X}$ are mapped onto a similarity space $\mathbf{X}_s$. In this space, $C$ centroids $\mathbf{X}_c$ are established uniformly, and their characteristics are distilled using an average pooling methodology. Subsequently, the cosine similarity matrix $\mathbf{S}$ between $\mathbf{X}_s$ and $\mathbf{X}_c$  is computed, resulting in $\mathbf{S} \in \mathbb{R}^{C \times N}$. Based on the similarity matrix 
$\mathbf{S}$, each vertex is then assigned to the nearest hyperedge. This process yields a primary set of hyperedges, denoted as $\mathcal{E}_c$.
\begin{figure}[t]
  \centering
    
    \includegraphics[width=\linewidth]{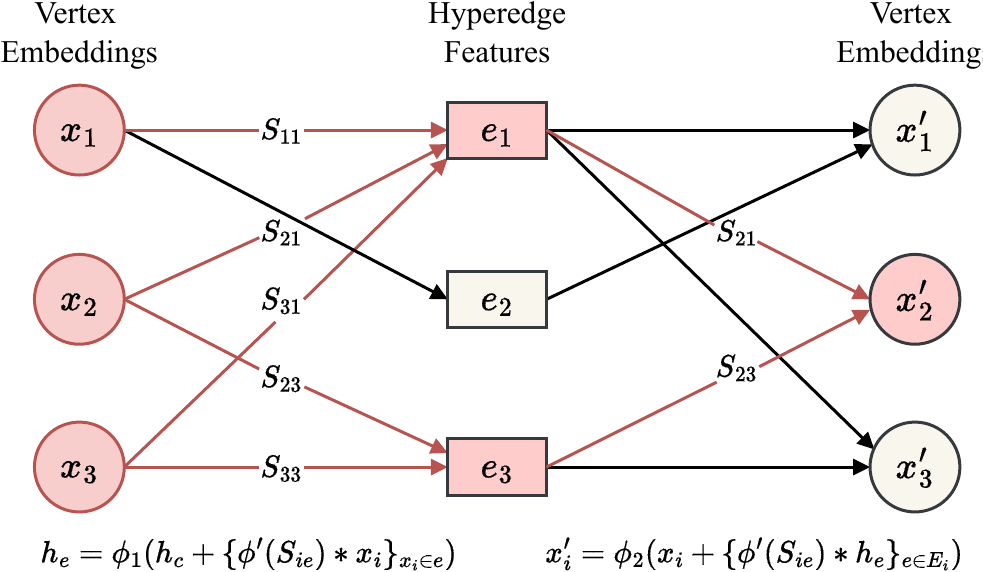}

  \caption{Illustration of two-stage message passing of our Dynamic Hypergrpah Convolution (DHConv). $\textbf{\textit{h}}_{c}$ is the feature of the hyperedge centroid, $\mathbf{S}_{ie}$ is the cosine similarity matrix between vertices and hyperedge centroids, and $\textbf{\textit{x}}_{i}$ and $\textbf{\textit{x}}'_{i}$ represent the vertex feature before and after DHConv. Note that how messages flow to vertex 2 is marked in red. }
\end{figure}

\noindent{\textbf{Region Partition. }}In terms of computational efficiency, the time complexity for comparing $N$ $D$-dimensional vectors across $C$ clusters is $O(NCD)$. To mitigate excessive computational demands, the image is partitioned into $m$ sub-regions akin to the Swin Transformer~\cite{swin}. This strategy reduces the time complexity to $O\left(\frac{NCD}{m}\right)$. While it may limit the scope of global information exchange, it introduces a beneficial inductive bias by promoting locality, akin to the sliding window approach observed in CNNs.

\noindent{\textbf{Dilated Hypergraph Construction. }}We propose a novel hypergraph construction strategy, named Dilated Hypergraph Construction (DHGC), to enhance the hyperedge set derived from clustering processing. In this approach, for each $ w \times w $ window, 
the central vertex $ v_c $ forms a series of dilated hyperedges. 
For the central vertex $ v_c $ in each window, we define a set of hyperedge 
neighborhoods $ N_k(v_c) $ for different dilation rates $ r \in {\mathbb{N}^+} $.
These hyperedge neighborhoods connect the central vertex in the window to the vertices
in its dilated neighborhood.

For a given dilation rate $ r $, the hyperedge neighborhood $ N_k(v_c) $ for each 
$ w \times w $ window's central vertex $ v_c $ includes $ K $ vertices, 
identified within a dilation step $ r $. The hyperedge comprises vertices 
whose coordinates, after dilation, remain inside the window's boundary.

The coordinates are defined as follows:
\begin{equation}
\begin{aligned}
    N_k(v_c) &= \left\{ (i', j') \middle| i' = i + p \times r, j' = j + q \times r \right\}, \\
    &\phantom{=} \left. -\frac{w}{2r} \leq p, q \leq \frac{w}{2r}, |N_k(v_c)| = K , \right.
\end{aligned}
\end{equation}
where $ (i, j) $ are the coordinates of $ v_c $, $ p, q $ are integers that scale 
with the dilation rate $ r $, and $ |N_k(v_c)| $ represents the number of vertices 
within the hyperedge structure. This arrangement ensures that each hyperedge captures 
a fixed number of vertices, enabling a consistent representation of spatial relationships 
across different scales.

Each dilated hyperedge is assigned to a unique sparsity-aware weight $ \mathbf{w}_r $, reflecting its specific dilation rate's contribution to the feature aggregation process. If there are 
$R$ dilation rates considered, the final weight matrix 
$ \mathbf{W} $ for the hypergraph is obtained by combining the $R$ individual hyperedge weights $ \mathbf{w}_r $, leading to:
\begin{equation}
    \mathbf{W} = \text{diag}(\mathbf{w}_1, \mathbf{w}_2, \cdots, \mathbf{w}_R).
\end{equation}

Consequently, our \emph{Dilated Hypergraph} is mathematically formulated as follows:
\begin{equation}
\mathcal{G}(\mathcal{V}, \mathcal{E},\mathbf{W},\mathbf{S}) = \left\{\mathcal{V}, \bigcup_{v_c \in \mathcal{V}} \mathcal{E}_{k,v_c} \cup \mathcal{E}_c, \mathbf{W}, \mathbf{S}\right\},
\end{equation}
where $ \mathcal{V} $ represents the entire set of vertices within the image or dataset, 
and $ \mathcal{E} $ is the cumulative union of all hyperedge neighborhoods derived from 
different dilation levels and all hyperedge sets formed by clustering, corresponding to 
every central vertex $ v_c $ within the windows. This structure not only enables the 
capture of the local information of an image but also facilitates the grasping of neighbor information 
at extended distances, according to varying dilation levels. By amalgamating features 
from different levels of dilation, we gain a deeper understanding of the intricate structure and interactions within an image.

\subsection{Dynamic 
Hypergraph Convolution}

In this subsection, we detail the architecture of our proposed Dynamic Hypergraph Convolution (DHConv), divided into two principal stages similar to~\cite{dhgnn, unignn}. The first stage, Vertex Convolution, aggregates vertex embeddings to constitute the hyperedge features. The second stage, Hyperedge Convolution, involves the distribution of these hyperedge features back to the associated vertices, which serves to update their individual embeddings.

\noindent{\textbf{Vertex Convolution.}}
In previous studies, vertex convolutions have relied on  techniques such as pooling, applying pre-defined transformation matrices based on graph structures, and dynamically learning the transformation matrix 
$\mathbf{T}$ utilizing MLPs and $1-d$ convolutions~\cite{dhgnn}.
Due to the computational complexity when dealing with vision tasks, the existing methods are unsuitable to handle the hypergraph-based vision data.

Our Dynamic Vertex Convolution approach is specifically designed to handle the high dimensionality and complexity of visual data. Unlike conventional methods, we leverage cosine similarity clustering hyperedges $ \mathcal{E}_c $, aggregating features adaptively through dynamic learning. 

Each given clustering hyperedge $ e $ consists of a hyperedge centroid $ h_c $ and multiple vertices $ x_i \in e $.  Our vertex feature aggregation process is formulated as follows:
\begin{equation}
\begin{gathered}
    h_e = \frac{1}{C} \left( h_c + \sum_{x_i\in e} \text{sig}(\alpha s_i + \beta) * x_i \right), \\
    \text{s.t.} \quad C = 1 + \sum_{x_i\in e} \text{sig}(\alpha s_i + \beta),
\end{gathered}
\end{equation}
where $ s_i $ represents the cosine similarity between vertex $ i $ and the hyperedge centroid $ h_c $, $ x_i $ are the feature vectors of vertex, and $ C $ is a normalization factor in ensuring numerical stability and emphasizing the locality in each hyperedge.

For each given dilated hyperedge 
$e$, the feature aggregation process is designed to capture multi-scale information efficiently. This process is formulated as follows:
\begin{equation}
    h_e = \frac{ h_c + \sum_{x_i\in e} w_r * x_i }{1 + |e| w_r},
\end{equation}
where $w_r$ is a learnable parameter, assigned to the dilated hyperedge $e$ with dilation $r$, reflecting the importance of features at different scales.

Through this approach, the proposed vertex convolution not only adapts to the complexity of hypergraph structure but also dynamically adjusts the interaction between vertices based on their cosine similarity to the hyperedge center. Thus, the proposed method can derive a  dynamic hypergraph image representation, which is more suitable for handling highly complex relationships  in different vision tasks.

\noindent{\textbf{Hyperedge Convolution.}} The proposed Hyperedge Convolution, inspired by the Graph Isomorphism Network (GIN) convolution~\cite{gin}, adaptively leverages hyperedge features to update the vertex embeddings based on cosine similarity or hyperedge weights, enabling feature exchange and fusion of local vertex information with hyperedge context.  The update formulation is defined as follows:
\begin{equation}
    z_i = \sum_{e \in E_i} (I_c(e) * \text{sig}(\alpha s_i + \beta) + I_d(e) * w_r) * h_e,
\end{equation}
\begin{equation}
    x'_{i} = FC\left(\sigma\left(\text{Conv}((1 + \varepsilon)x_i + 
    z_i)\right)\right),
\end{equation}
 where $ h_e $ denotes the aggregated feature of hyperedge $e$, $ E_i $ denotes a set of all hyperedges connected to vertex $i$,  $ \varepsilon $ is a learnable parameter,  $ \sigma(\cdot)$ is a non-linear activation function, $FC$ is a fully connected layer, $I_c(e)$ and $I_d(e)$ are the indicator functions for the two types of hyperedges, with $I_c(e), I_d(e) \in \{0,1\}$ and $I_c(e) \neq I_d(e)$.

 \begin{table}[t]
	\small 
	\centering
    
	\renewcommand{\arraystretch}{1.0}
	\setlength{\tabcolsep}{2pt}{
		\begin{tabular}{l|c|c|c|c} 
			\toprule[1.0pt]
   
			 Model & Type & Params (M) & FLOPs (G)  & Top-1 (\%)\\

			\midrule
    ResNet-18~~\cite{resnet} & CNN & 12.0 & 1.8 & 70.6  \\ 
    ResNet-50~~\cite{resnet} & CNN & 25.6 & 4.1 & 79.8  \\
    InceptionNeXt-T~\cite{inceptionnext} & CNN  &28.0 & 4.2 & 82.3  \\ 
    InceptionNeXt-S~\cite{inceptionnext}& CNN  & 49.0 & 8.4 & 83.5  \\ 
    InceptionNeXt-B~\cite{inceptionnext}& CNN  & 87.0 & 14.9 & 84.0  \\ 
    \midrule
    Swin-T~\cite{swin} & ViT & 29.0 & 4.5 & 81.3 \\
    Swin-S~\cite{swin} & ViT & 50.0 & 8.7 & 83.0 \\
    Swin-B~\cite{swin} & ViT & 88.0 & 15.4 & 83.5 \\
    PVTv2-B1~~\cite{pvtv2} & ViT  & 13.1 & 2.1 & 78.7  \\
    PVTv2-B2~~\cite{pvtv2} & ViT & 25.4 & 4.0 & 82.0  \\
    FLatten-Swin-T~~\cite{flattenvit} & ViT&  29.0 & 4.5 & 82.5  \\
    FLatten-Swin-S~~\cite{flattenvit} & ViT&  51.0 & 8.7 & 83.5  \\
    FLatten-Swin-B~~\cite{flattenvit} & ViT&  89.0 & 15.4 & 83.8  \\
    \midrule
    Poolformer-S12~\cite{metaformer} & Pool & 12.0 & 2.0 & 77.2  \\ 
    Poolformer-S36~\cite{metaformer} & Pool & 31.0 & 11.2 & 80.3  \\
    Poolformer-M48~\cite{metaformer} & Pool & 73.0 & 15.9 & 81.4  \\
    \midrule
    Vim-Tiny~~\cite{vim} & SSM& 7.0 & - & 76.1  \\ 
    Vim-Small~~\cite{vim} & SSM& 26.0 & 5.1 & 80.5  \\ 
    Vim-Base~~\cite{vim} & SSM& 98.0 & - & 81.9  \\ 
    VMamba-T~~\cite{vmamba} & SSM& 22.0 & 5.6 & 82.2  \\ 
    VMamba-M~~\cite{vmamba} & SSM& 44.0 & 11.2 & 83.5  \\
    VMamba-B~~\cite{vmamba} & SSM& 75.0 & 18.0 & 83.7  \\
    \midrule
     
    CoC-S~~\cite{Context-Cluster} &Cluster &  14.0 & 2.6  & 77.5 \\
    CoC-M~~\cite{Context-Cluster} &Cluster& 27.9 & 5.5  & 81.0 \\
    ClusterFormer-T~~\cite{clusterformer}& Cluster& 28.0 & 5.7 & 81.5  \\
    ClusterFormer-S~~\cite{clusterformer}& Cluster& 48.7 & 9.9 & 83.4  \\
    FEC-Small~~\cite{fec}& Cluster&  5.5 & 1.4 & 72.7  \\
    FEC-Base~~\cite{fec}& Cluster&  14.4 & 3.4 & 78.1  \\
    FEC-Large~~\cite{fec}& Cluster&  28.3 & 6.5 & 81.2  \\
    \midrule
    Pyramid ViG-Ti~~\cite{vig} & GNN & 10.7 & 1.7 & 78.2 \\	
    Pyramid ViG-S~~\cite{vig} & GNN  & 27.3 & 4.6 & 82.1  \\
    Pyramid ViG-M~~\cite{vig} & GNN  & 52.4 & 8.9 & 83.1  \\
    Pyramid ViG-B~~\cite{vig} & GNN  & 92.6 & 16.8 & 83.7  \\
    \midrule
    PViHGNN-Ti ~~\cite{vihgnn} & HGNN & 12.3 & 2.3 & 78.9 \\
    PViHGNN-S ~~\cite{vihgnn} & HGNN & 28.5 & 6.3 & 82.5 \\
    PViHGNN-M ~~\cite{vihgnn} & HGNN & 52.4 & 10.7 & 83.4 \\
    PViHGNN-B ~~\cite{vihgnn} & HGNN & 94.4 & 18.1 & 83.9 \\
    \midrule 
    \cellcolor{mylb}DVHGNN-T (ours) & \cellcolor{mylb}HGNN & \cellcolor{mylb}11.1 &\cellcolor{mylb}1.9 & \cellcolor{mylb}\textbf{79.8} \\
     \cellcolor{mylb}DVHGNN-S (ours) & \cellcolor{mylb}HGNN & \cellcolor{mylb}30.2  & \cellcolor{mylb}5.2 &  \cellcolor{mylb}\textbf{83.1} \\
      \cellcolor{mylb}DVHGNN-M (ours) & \cellcolor{mylb}HGNN & \cellcolor{mylb}52.5 & \cellcolor{mylb}10.4 &  \cellcolor{mylb}\textbf{83.8} \\
     \cellcolor{mylb}DVHGNN-B (ours) & \cellcolor{mylb}HGNN &\cellcolor{mylb}92.8 & \cellcolor{mylb}16.8 &  \cellcolor{mylb}\textbf{84.2} \\
			\bottomrule[1pt]
		\end{tabular}
	}
 \caption{Results of DVHGNN variants and other backbones on ImageNet-1K. All the models are trained at 224 $\times$ 224 resolution}
 \label{tab:classfication}
\end{table}

\subsection{Multi-head Computation}
Multi-head computations have been extensively studied in transformer architecture~\cite{transfomer}. Our DVHGNN model continues to implement these computations and update mechanisms as described in~\cite{Context-Cluster}. Specifically, we employ $h$ heads, standardizing the dimensions of the value space $X_v$ and similarity space $X_s$ to $D'$ for simplicity. The outputs of the multi-head operations are combined and integrated via a FC layer, resulting in $h$ distinct hypergraphs. This multi-head computation allows the model to concurrently update features across multiple representational hypergraph subspaces, thereby enhancing feature diversity.

\subsection{Dilated Vision HGNN Architecture}

As shown in Figure. \ref{fig:attn_comparison}, the basic component of our proposed architecture is the Multi-Scale (MS) DVHGNN block. Each MS DVHGNN block consists of three modules: MS Hypergraph Construction, DHConv, and ConvFFN. To make the model compatible with various downstream vision tasks, we adopt a four-stage pyramid structure. Within each stage, a convolutional layer is employed to reduce the dimensions of the input feature map to a quarter of its original height and width
 $ (\frac{H}{4} \times \frac{W}{4})$, which is subsequently succeeded by a series of MS ViHGNN blocks. Culminating the architecture, a prediction head is deployed for image classification. Detailed configurations of our DVHGNN variants are listed in appendices, with $D'$ refers to the dimension of each head.

\section{Experiments}
\subsection{Image Classification on ImageNet}
\noindent{\textbf{Experimental Settings:}} 
We benchmark our models on ImageNet-1K~\cite{imagenet}, a dataset with 1.3 million training images and 50,000 validation images across 1,000 classes. The training was conducted over 300 epochs with an image resolution of 224$\times$224, using AdamW~\cite{AdamW} optimizer. We enhanced our models' generalization with a combination of data augmentation and regularization techniques, including RandAugment~\cite{randaugment}, Mixup~\cite{mixup}, CutMix~\cite{cutmix}, and Random Erasing~\cite{randomeras}, with weight decay, Label Smoothing~\cite{labelsmooth} and Stochastic Depth~\cite{stochasticpath}. Our implementation leverages PyTorch~\cite{pytorch} and Timm~\cite{timm}. 

\noindent{\textbf{Experimental Results:}} As depicted in Table.~\ref{tab:classfication}, DVHGNN models outperform other state-of-the-art vision backbones, especially the GNN-based and HGNN-based models. Specifically, DVHGNN-S achieves a Top-1 accuracy of 83.1\%, surpassing ViG-S by 1.0\% and ViHGNN by 0.6\% with 18\% less FLOPs. These results highlight our model's ability to capture complex visual representations effectively with increased computational efficiency.

\begin{table}[t]
    \small
    \centering
 \setlength\tabcolsep{0.5pt}
	\begin{tabular}{l|cc|ccc|ccc}
		\toprule[1pt]
		\multirow{2}{*}{Backbone} & \multicolumn{8}{c}{RetinaNet 1$\times$}  \\  
		\cline{2-9} 
		& Param & FLOPs & AP & AP$_{50}$ & AP$_{75}$  & AP$_{\rm S}$ & AP$_{\rm M}$ &  AP$_{\rm L}$ \\
		\midrule		
		ResNet-50~\cite{resnet} & 38M & 239G & 36.3 & 55.3 & 38.6 & 19.3 & 40.0 & 48.8 \\

		PVT-Small~\cite{pvt} & 34M & 227G & 40.4 & 61.3 & 44.2 & 25.0 & 42.9 & 55.7 \\

            Swin-T~\cite{swin} & 39M & 245G & 41.5 & 62.1 & 44.2 & 25.1 & 44.9 & 55.5 \\
  		Slide-PVT-S~\cite{SlideTransformer} & - & 251G & 42.4 & 63.9 & 45.0 & 26.8 & 45.6& 56.9 \\
		
		Pyramid ViG-S~\cite{vig} & 36M & 240G & 41.8 & 63.1 & 44.7 & 28.5 & 45.4 & 53.4 \\
  
    			PViHGNN-S~\cite{vihgnn} & 38M & 244G & 42.2 & 63.8 & 45.1 & \textbf{29.3} & 45.9 & \textbf{55.7} \\	
           \br DVHGNN-S (ours) & 38M & 242G  & \textbf{43.3} &\textbf{64.3} & \textbf{46.3} & 28.3  & \textbf{47.9}  & 54.6  \\
		\midrule
		\multirow{2}{*}{Backbone}  &\multicolumn{8}{c}{Mask R-CNN 1$\times$} \\
		\cline{2-9}
		& Param & FLOPs & AP$^{\rm b}$ & AP$_{50}^{\rm b}$ & AP$_{75}^{\rm b}$  & AP$^{\rm m}$ & AP$_{50}^{\rm m}$ & AP$_{75}^{\rm m}$ \\
        \midrule

		ResNet-50~\cite{resnet} & 44M  & 260G & 38.0 & 58.6 & 41.4 & 34.4 & 55.1 & 36.7 \\
		PVT-Small~\cite{pvt} & 44M & 245G & 40.4 & 62.9 & 43.8 & 37.8 & 60.1 & 40.3 \\
          Swin-T~\cite{swin} & 48M & 264G & 42.2 & 64.6 & 46.2 & 39.1 & 61.6 & 42.0  \\
   		Slide-PVT-S~\cite{SlideTransformer} & 42M & 269G & 42.8 & 65.9 & 46.7 & 40.1 & 63.1& 43.1 \\
     ConvNeXt-T~\cite{convnet} & 48M & 262G & 44.2 & 66.6 & 48.3 & 40.1 & 63.3 & 42.8 \\


  		Pyramid ViG-S~\cite{vig} & 46M & 259G & 42.6 & 65.2 & {46.0} & 39.4 & 62.4 & {41.6} \\
      			PViHGNN-S~\cite{vihgnn} & 48M & 262G  & 43.1 & 66.0 & 46.5 & 39.6 & 63.0 & 42.3 \\

       \br DVHGNN-S (ours) & 49M & 261G & \textbf{44.8} & \textbf{66.8} & \textbf{49.0} & \textbf{40.2} & \textbf{63.5} & \textbf{43.1} \\
		\bottomrule[1pt]
	\end{tabular}
\caption{Results of object detection and instance segmentation on COCO 2017. The input size is 1280 $\times$ 800.}
\label{table:cocodet}

\end{table}
 
\subsection{Object Detection and Instance Segmentation}
\noindent{\textbf{Experimental Settings:}}
The experiments were conducted on the MS-COCO~\cite{coco} dataset, which includes 118K training images, validation images, and 20K test images. Respectively, for objection detection and instance segmentation, We employed DVHGNN-S pre-trained on ImageNet-1K as a backbone and incorporated it into two detectors: RetinaNet~\cite{retinanet} and Mask R-CNN~\cite{maskrcnn}. Following common practice, we trained the two downstream vision tasks with DVHGNN-S for 12 (1 $\times$ schedule) epochs and the implementation was performed by MMDetection~\cite{mmdetection}.

\noindent{\textbf{Experimental Results:}} As depicted in Table.~\ref{table:cocodet}, our DVHGNN outperforms the state-of-the-art backbones. Specifically, under the RetinaNet framework, our DVHGNN-S achieves 43.3\% mAP that surpasses ViG by 1.5\% and ViHGNN by 1.1\%, respectively. Under the Mask R-CNN framework, our DVHGNN-S achieves 44.8\% bbox mAP and 40.2\% mask mAP that surpasses ViHGNN by 1.7\% and 0.6\%, respectively. 

\subsection{Semantic Segmentation on ADE20K}
\noindent{\textbf{Experimental Settings:}} We evaluated the semantic segmentation performance of our DVHGNN on ADE20K~\cite{ade20k} dataset using two representative frameworks: Upernet~\cite{upernet} and Semantic FPN~\cite{SemanticFPN}, with our pre-trained DVHGNN-S as the backbone. For Upernet, we followed the Swin Transformer~\cite{swin} configuration and trained our model for 160K iterations. As for Semantic FPN with 80K iterations, we followed the PVT~\cite{pvt} configuration.

\noindent{\textbf{Experimental Results:}} 
Table.~\ref{tab:semantic_seg} shows the superior results of our DVHGNN with other popular backbones on the ADE20K validation set. Specifically, using the Semantic FPN~\cite{SemanticFPN} framework, our DVHGNN-S achieves 43.8\% mIoU, surpassing Swin-T~\cite{swin} by 2.3 \%. Under UperNet~\cite{upernet} framework, our DVHGNN-S achieves 46.8\% mIoU, surparssing Swin-T~\cite{swin} by 2.3 \%.
\begin{table}[t]
    \small
    \centering
    
    \setlength\tabcolsep{5pt}
    \begin{tabular}{l|c|c|c|c}
        \toprule[1pt]
        \multirow{2}{*}{Backbone} & \multirow{2}{*}{Method} & Params & FLOPs & mIoU \\
        & & (M) & (G) & (\%)  \\
        \midrule
        ResNet-50~\cite{resnet} & S-FPN & 29 & 183 & 36.7 \\
        Swin-T \cite{swin} & S-FPN & 32 & 182 & 41.5 \\
        PVT-S \cite{pvt} & S-FPN & 28 & 225 & 42.0 \\
        Slide-PVT-S \cite{SlideTransformer} & S-FPN & 26 & 188 & 42.5 \\
        DAT-T \cite{dat} & S-FPN & 32 & 198 & 42.6 \\
        InceptionNeXt-T \cite{inceptionnext} & S-FPN & 28 & - & 43.1 \\
        \br
        DVHGNN-S (ours) & S-FPN & 32 & 181 & \textbf{43.8} \\
        \hline
        Swin-T \cite{swin} & UperNet & 60 & 945 & 44.5 \\
        FLaTeN-Swin-T~\cite{flattenvit}  & UperNet & 60 & 946 & 44.8 \\
        DAT-T \cite{dat} & UperNet & 60 & 957 & 45.5 \\
        Slide-Swin-T \cite{SlideTransformer} & UperNet & 60 & 946 & 45.7 \\
        ConvNeXt-T \cite{convnet} & UperNet & 60 & 945 & 46.1 \\
        \br
        DVHGNN-S (ours) & UperNet & 60 & 945 & \textbf{46.8} \\
         \bottomrule[1pt]
    \end{tabular}
    
    \caption{ Results of semantic segmentation on ADE20K validation set. The input size is 2048 $\times$ 512.}
    \label{tab:semantic_seg}
\end{table}
\begin{table}[t]

    \small
    \centering


\setlength{\tabcolsep}{5pt}{
\begin{tabular}{c|c|c|c|c}
\toprule[1pt]
\multirow{2}{*}{Step} & \multirow{2}{*}{Method} & Params &  FLOPs & Top-1 acc \\
& & (M) & (G) & (\%) \\
\midrule
0 & Feature Dispatching & 7.7  & 2.1   & 76.1  \\
1 & More Heads \& Thinner & 11.0  & 1.7   & 76.1  \\
2 & DHConv & 11.1  & 1.8   & 77.2\textsubscript{\textcolor{red}{(+1.1)}}  \\
3 & +DHGC & 11.1  & 1.8 & 78.0\textsubscript{\textcolor{red}{(+0.8)}} \\
4 & ++ConvFFN & 11.2 & 1.9 & 78.4\textsubscript{\textcolor{red}{(+0.4)}} \\
\bottomrule[1pt]
\end{tabular}

}

\caption{Ablation experiments of DVHGNN, where ``Feature Dipatching", ``DHConv" and ``DHGC"epresent Feature Dipatching in baseline model, Dynamic Hypergraph Convolution, Dilated Hypergraph Construction. The training epochs is set to 250.}
\label{table:ablation1}
\end{table}

\begin{figure*}[t]

	\centering
	\small
	\setlength{\tabcolsep}{8pt}{
		\begin{tabular}{cccc}
			\makecell*[c]{\includegraphics[width=0.22\linewidth]{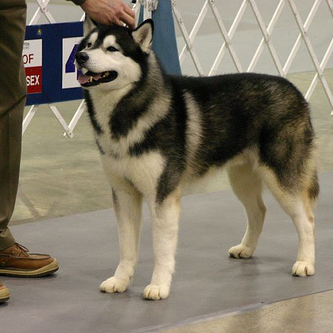}}  &
			\makecell*[c]{\includegraphics[width=0.22\linewidth]{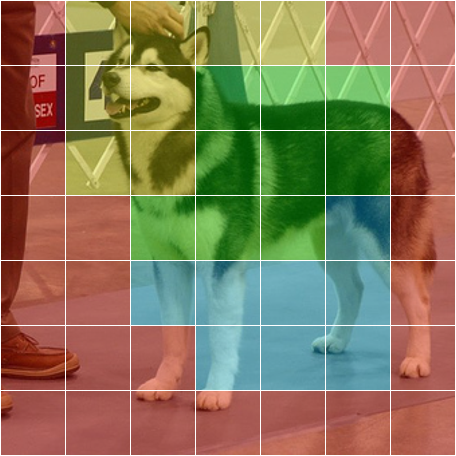}}  & 
            \makecell*[c]{\includegraphics[width=0.22\linewidth]{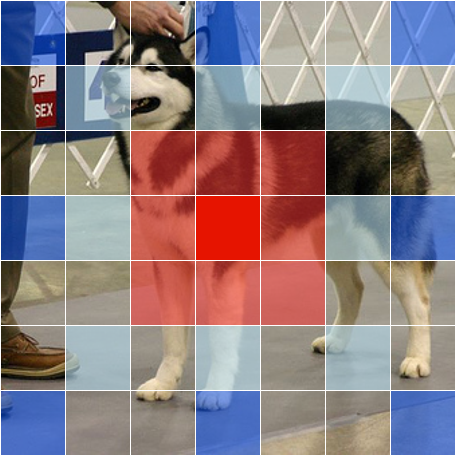}} &
			\makecell*[c]{\includegraphics[width=0.22\linewidth]{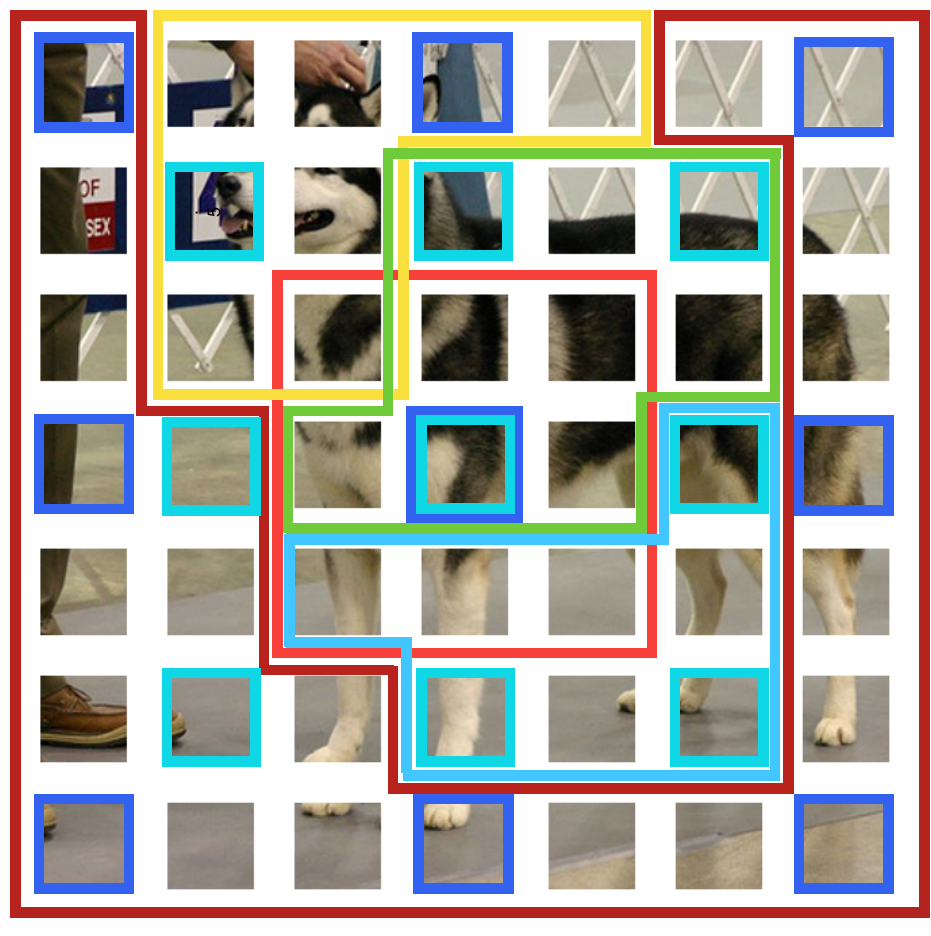}}  
			\\
			(a) Input image. & (b) Hyperedges of Clustering. & (c) Hyperedges of DHGC. & (d) Hypergraph structure.
		\end{tabular}
	}
	\caption{Visualization of the hypergraph structure of DVHGNN. The  hypergraph structure is obtained by an overlay of hyperedges derived from the Clustering method and DHGC.}
 \label{Fig:Visualization1}
\end{figure*}
\subsection{Ablation Study}
\noindent \textbf{Ablation study of modules. }In our structured ablation experiments on the ImageNet-1K dataset, as depicted in Table.~\ref{table:ablation1}, we trained the DVHGNN-T model for 250 epochs to systematically evaluate the contributions of its modules. In Step 0, we aligned the channel configuration and downsampling stages of the baseline model, ContextCluster-Ti~\cite{Context-Cluster}, with those of DVHGNN-T, achieving a Top-1 accuracy of 76.1\%. In Step 1, we increased the number of heads while making the model thinner and deeper, which reduced the computational cost by 19\% GFLOPs. In Step 2, the integration of our Dynamic Hypergraph Convolution, replacing the standard Feature Dispatching mechanism, led to a 0.9\% improvement in Top-1 accuracy. Step 3 involved the incorporation of dilated hyperedges via DHGC, further boosting accuracy by 0.8\%. Finally, in Step 4, replacing the conventional MLP with the ConvFFN architecture resulted in an additional 0.3\% increase in Top-1 accuracy. 

\noindent \textbf{Ablation Study of Hyperparameters.} To rigorously assess the impact of key hyperparameters on DVHGNN performance, we conducted comprehensive ablation studies on four crucial factors: (1) the number of clustering centroids $C$, (2) the number of attention heads per stage $h$, (3) the feature dimension of each head $D'$, and (4) the minimum kernel size (dilation factors). All experiments were conducted using the DVHGNN-T variant, trained on ImageNet-1K. As depicted in Table.~\ref{table:ablation_C}, increasing the number of clustering centroids beyond a certain threshold leads to a marginal decline in Top-1 accuracy, likely due to redundancy and over-clustering, which degrades feature aggregation. Thus, we adopt $C=4$ as the optimal setting. Conversely, Table.~\ref{table:ablation_h} and Table.~\ref{table:ablation_dim} demonstrate that increasing the number of attention heads $h$ and the feature dimension per head $D'$ consistently enhances performance, underscoring the benefits of richer representation learning. Moreover, as indicated in Table.~\ref{table:ablation_kernel}, employing a 7$\times$7 minimum kernel size yields the best results, emphasizing the advantages of larger receptive fields for spatial feature extraction.

\begin{table}[t]

\centering
\small

\setlength{\tabcolsep}{7pt}{
\begin{tabular}{c|c|c|c|c}
\toprule[1pt]
$C$  & Params (M) &  FLOPs (G) & Top-1 (\%) & Top-5 (\%)\\
\midrule
4 & 11.2 & 1.9  & \textbf{79.8}   & \textbf{95.3}  \\
9 & 11.3 & 1.95  & 79.6   & 95.0  \\
16 & 11.5 & 2.0 & 79.5   & 94.9  \\
25 & 13.0 & 2.1 & 79.7   & 95.1  \\
\bottomrule[1pt]
\end{tabular}}
\caption{Effects of number of clustering centroids $C$ (300 epochs).}

\label{table:ablation_C}
\end{table}
\begin{table}[t]

\centering
\small

\setlength{\tabcolsep}{7pt}{
\begin{tabular}{c|c|c|c}
\toprule[1pt]
$h$  & Params (M) &  FLOPs (G)  & Top-1 (\%) \\
\midrule
\text{[2, 4, 8, 16]} & 10.2 & 1.76 & 74.8   \\

\text{[3, 6, 12, 24]} & 11.2 & 1.92  & 75.8   \\
\text{[4, 8, 16, 32]} & 12.2 & 2.08 & \textbf{76.3}   \\

\bottomrule[1pt]
\end{tabular}}
\caption{Effects of number of heads $h$ (200 epochs). }
\label{table:ablation_h}
\end{table}
\subsection{Visualization}
To provide a more intuitive evaluation of our proposed DVHGNN model, we visualize the hypergraph construction of one head in the DVHGNN-S model after the last stage. As depicted in Figure.~\ref{Fig:Visualization1}, the background areas on the left, right, and bottom are grouped into one hyperedge (represented by red patches), while distinct components of the dog, including its head, body, and limbs, are assigned to separate hyperedges (represented by yellow, green, and blue patches, respectively). This observation highlights that our model not only effectively captures global semantic structures but also comprehends intricate intra-class relationships among object parts, demonstrates the strength of DVHGNN in learning high-level semantic information, enabling accurate and structured object understanding. 
\begin{table}[t]

\centering
\small

\setlength{\tabcolsep}{7pt}{
\begin{tabular}{c|c|c|c|c}
\toprule[1pt]
$D'$  & Params (M) &  FLOPs (G)  & Top-1 (\%) & Top-5 (\%) \\
\midrule
16 & 10.4 & 1.75  & 75.3 & 92.6   \\
24 & 11.2 & 1.92 & 75.8  & 93.1  \\
32 & 12.0 & 2.10 & \textbf{76.1} & \textbf{93.3}   \\
\bottomrule[1pt]
\end{tabular}}
\caption{Effects of dimension of each head $D'$ (200 epochs). }
\label{table:ablation_dim}
\end{table}
\begin{table}[t]

\centering
\small

\setlength{\tabcolsep}{5pt}{
\begin{tabular}{c|c|c|c}
\toprule[1pt]
Minimum Kernel size  & Params (M) &  FLOPs (G) & Top-1 (\%) \\
\midrule
3 $\times$ 3 & 11.2 & 1.87  & 75.5   \\
5 $\times$ 5 & 11.2 & 1.89 & 75.6   \\
7 $\times$ 7 & 11.2 & 1.92 & \textbf{75.8}    \\
\bottomrule[1pt]
\end{tabular}}
\caption{Effects of different Mininum Kernel sizes (200 epochs). }
\label{table:ablation_kernel}
\end{table}
\section{Conclusion}

In this paper, we have proposed  a novel vision backbone architecture, termed the DVHGNN, to learn hypergraph-aware vision features. The proposed method is characterized by its dynamic and learnable hypergraph, which can adaptively capture the multi-scale dependencies of an image. Furthermore, a novel dynamic hypergraph convolution is designed to aggregate vertex features into hyperedge features. The extensive qualitative and quantitative experimental results on the benchmark vision datasets demonstrate that the proposed DVHGNN significantly enhances the learning performance of different vision tasks, which  achieves remarkable Top-1 accuracy on the ImageNet-1K. In future work, we will  further explore the scalability and generalization of the proposed across the other vision tasks.

\clearpage
\noindent \textbf{Acknowledgments }The work was supported by the National Science Fund for Distinguished Young Scholars (No.62025603) and National Natural Science Foundation of China (No. 62072386), supported by Yunnan Provincial Major S\&T Special Plan Project (No. 202402AD080001), Henan Key R\&D Project (No. 231111212000), Henan Center for Outstanding Overseas Scientists (No. GZS2022011), Open Foundation of Henan Key Lab of General Aviation Technology (No. ZHKF-230212), Key Lab of Oracle Information Processing of MOE (No. OIP2024E002). We also thank Digital Culture Lab of Tencent SSV for their invaluable contribution during our research.

{
    \small
    \bibliographystyle{ieeenat_fullname}
    \bibliography{main}
}

\clearpage
\setcounter{page}{1}
\maketitlesupplementary
\renewcommand{\thesection}
{\Alph{section}}
\appendix

\section{Cross-task Visualizations}
\begin{figure}[t]
    \centering
     
     \includegraphics[width=1.0\linewidth]{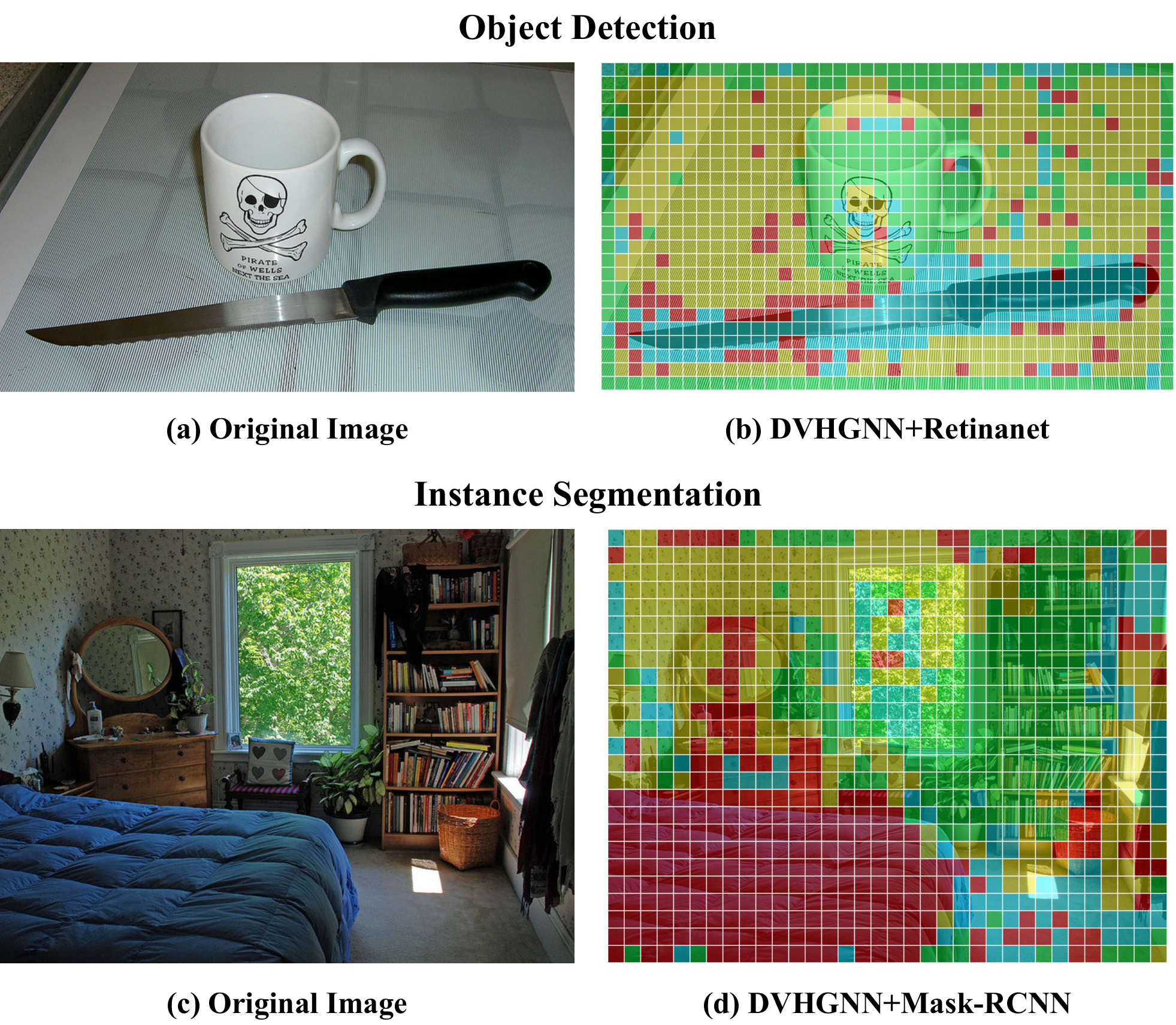}

    \caption{Visualization of cross-tasks clustering hyperedges in object detection and instance segmentation, respectively.}

    \label{fig:visul}
\end{figure}

In Figure~\ref{fig:visul}, we present a visualization of the clustering hyperedges extracted from a single attention head in the final stage of DVHGNN-S, where the feature map size is set to 25$\times$38. Specifically, Figure~\ref{fig:visul} (b) and Figure~\ref{fig:visul} (d) illustrate the clustering results for object detection and instance segmentation tasks, respectively, under the RetinaNet and Mask R-CNN architectures. To facilitate interpretation, different hyperedge types are distinguished by color, encoding the underlying semantic relationships among visual entities. For example, in Figure~\ref{fig:visul} (b), the blue hyperedges correspond to the knife category, while in Figure~\ref{fig:visul} (d), the red hyperedges represent the bed category. Notably, we omit the visualization of dilated hyperedges, as their structural configurations remain consistent with those elaborated in the main paper. This visualization further substantiates the efficacy of our hypergraph-based approach in capturing meaningful semantic structures across diverse tasks.
\section{Additional Ablation Study}
\noindent \textbf{Impact of the Clustering Step on Inference Speed.} Since the clustering method constitutes the fundamental backbone of our approach, direct ablation analysis on this component is not feasible. Therefore, we instead assess the impact of the proposed Dynamic Hypergraph Convolution (DHGC) and Dynamic Hypergraph Convolutional Filtering (DHConv) on inference efficiency. As reported in Table~\ref{table:ablation_infer}, removing DHGC leads to an inference speed improvement of 0.084 ms per image, while removing DHConv further accelerates inference by 0.087 ms per image. However, these speed gains come at the cost of a noticeable degradation in Top-1 accuracy, with DHGC and DHConv ablations resulting in 0.8\% and 1.1\% drops in performance, respectively. These results underscore the inherent trade-off between computational efficiency and predictive accuracy. While eliminating these components marginally improves throughput, the corresponding accuracy degradation suggests that DHGC and DHConv play a crucial role in enhancing model representation capacity and overall performance. Therefore, the slight increase in computational overhead is justified by the substantial accuracy gains, reinforcing the effectiveness of our proposed hypergraph-based feature aggregation strategy.

\begin{table}[t]

    \small
    \centering


\setlength{\tabcolsep}{3pt}{
\begin{tabular}{c|c|c|c|c}
\toprule[1pt]
 Method &  Params &  FLOPs & Throughput & Top-1\\
\midrule
 DVHGNN-T\textsubscript{w/o ConvFFN} & 11.1 M  & 1.8 G & 874.5 img/ms & 78.0\%   \\
 w/o DHGC & 11.1 M  & 1.8 G   & 943.8 img/ms & 77.2\% \\
 w/o DHConv & 11.0 M & 1.7 G   & 1028.1 img/ms & 76.1\% \\
\bottomrule[1pt]
\end{tabular}

}

\caption{Impact of distinct moudule on inference throughput.}

\label{table:ablation_infer}
\end{table}
\begin{table}[ht]
     \small
     \centering
     
     \setlength{\tabcolsep}{1.8pt}{
     \begin{tabular}{c|c|c|c|c}
        \toprule[1pt]
          Models & Channels & Blocks & Heads & $D'$  \\
          \midrule
          DVHGNN-T & [48, 96, 240, 480] & [2, 2, 6, 2] & [3, 6, 12, 24] & 24 \\ \midrule
          DVHGNN-S & [64, 128, 320, 640] & [3, 3, 9, 3] & [4, 8, 16, 32] & 32 \\ \midrule
          DVHGNN-M & [96, 192, 384, 768] & [4, 4, 14, 4] & [4, 8, 16, 32] & 32 \\ \midrule
          DVHGNN-T & [96, 192, 384, 768] & [6, 6, 24, 6] & [5, 10, 20, 40] & 32 \\
        \bottomrule[1pt]
     \end{tabular}
     }
     \caption{Configurations of different DVHGNN variants. }
     \label{tab:configurations}
 \end{table}

\end{document}